%% file: article.tex
\documentclass{article}
\usepackage{subfigure}
\usepackage{llncsdoc}
\usepackage{times} 
\usepackage{latexsym}
\usepackage{url}
\usepackage{graphicx}
\usepackage{latexsym}
\usepackage{amsmath, amsthm, amssymb, amsthm}
\usepackage{algorithm}
\usepackage{multirow}
\usepackage{array}
\usepackage{authblk}
\usepackage[noend]{algorithmic}

\newtheorem{theorem}{Theorem}
\newtheorem{lemma}{Lemma}
\newtheorem{proposition}{Proposition}

\newtheorem{definition}{Definition}

\begin{document}

\title{Symmetry-Based Search Space Reduction For Grid Maps}
\author{Daniel Harabor}
\author{Adi Botea}
\author{Philip Kilby}
\affil{NICTA and The Australian National University \\
Email: firstname.lastname@nicta.com.au}

\maketitle

\input abstract
\input introduction

\input relatedwork
\input empty_rooms

\input eight_connected

\input branching_factor

\input memory
\input setup
\input results

\input conclusion
\input acknowledgements

\bibliographystyle{splncs03}
\bibliography{references}
\end{document}

%% file: abstract.tex
\begin{abstract}
In this paper we explore a symmetry-based search space reduction technique
 which can speed up optimal pathfinding on undirected uniform-cost grid maps by
 up to 38
times.  Our technique decomposes grid maps into a set of empty rectangles,
removing from each rectangle all interior nodes and possibly some from along the
perimeter. We then add a series of macro-edges between selected pairs of
remaining perimeter nodes to facilitate provably optimal traversal through each
rectangle.  We also develop a novel online pruning technique to further speed up
search.  Our algorithm is fast, memory efficient and retains the same optimality
and completeness guarantees as searching on an unmodified grid map.
\end{abstract}

%% file: introduction.tex
\section{Introduction}
\label{sec:introduction}

Pathfinding on uniform-cost undirected grid maps is a problem commonly appearing in
the literature: for example in application areas such as robotics \cite{lee09}
artificial intelligence \cite{wang09} and video games \cite{davis00,sturtevant10}.  
In such contexts it is often the
case that queries sent to the pathfinding system need to be solved as quickly as
possible.  Traditionally, this requirement is met through the application of
hierarchical decomposition techniques that transform the search space into a
much smaller approximate representation \cite{botea04,sturtevant10}.
Such methods are very fast, particularly when compared to the classical A*
algorithm, but have the disadvantage that solutions found in the abstract state
space are often not optimal when mapped back to the original grid.  An
alternative speedup method is to develop better heuristics to guide the
search~\cite{bjornsson06,sturtevant09,goldenberg10}.  Though usually
fast, optimal and more effective than the popular Manhattan or Octile heuristic (both
analogous to Euclidean distance but optimised to 4 and 8-connected grids), they
have the disadvantage of requiring significant memory overhead.
\par
In this paper we present Rectangular Symmetry Reduction (RSR): a graph pruning
algorithm for undirected uniform-cost grid maps which is fast, memory efficient,
optimality preserving and which can, in some cases, eliminate entirely the need
to search.  The central idea that we will explore involves the identification
and elimination of path symmetries from the search space. 

To deal with path symmetries RSR makes use of an off-line
empty rectangle decomposition \cite{harabor10} that converts an arbitrary
undirected uniform-cost grid map into an equivalent one where only nodes
from the perimeter of each empty rectangle need to be explored during search.
We extend this approach in several directions: (i) we generalise the method
from 4-connected grid maps to the 8-connected case where the increase in
branching factor makes effective symmetry elimination more challenging; (ii) we
develop a new offline pruning technique that reduces the number of nodes which
need to be explored during search; (iii) we give a novel online pruning strategy
which speeds up node expansion by selectively evaluating either all neighbours
associated with a particular node or only a small subset.  We prove that in each
case both optimality and completeness are preserved.
\par
We perform a thorough empirical analysis, comparing RSR with three similar
state-of-the-art graph pruning algorithms on a
number of synthetic and realistic benchmarks, including one well known set from
the popular roleplaying game \emph{Baldur's Gate II}.  Compared to Harabor and
Botea's method \cite{harabor10}, we both extend the applicability and improve
the speed on the subset of instances where both algorithms are applicable.
We also compare RSR to the recent Swamps-based pruning method of Pochter \emph{et
al}~\cite{pochter10} and the enhanced Portal Heuristic of Goldenberg \emph{et
al}~\cite{goldenberg10}. 
We show that RSR has complementary strengths compared to both of these methods 
and identify classes of instances where RSR is clearly the better choice, 
dominating convincingly across a large number of instances.

%% file: relatedwork.tex
\section{Related Work}
\label{sec:relatedwork}
In the presence of symmetry, search algorithms often
evaluate many equivalent states and make little real progress toward the goal.
The topic of how best to deal with symmetry has received significant attention
in other parts of the literature~\cite{rossi06} but there are very few works
that explicitly identify and deal with symmetry in pathfinding domains 
such as grid maps. The only work of which we are aware is Empty Rectangular Rooms
\cite{harabor10}: a symmetry breaking technique specific to 4-connected
uniform-cost grid maps which we refer to as 4ERR.  We discuss the main
differences between 4ERR and RSR in Sections \ref{sec:introduction} and
\ref{sec:rsr}.
\par
The \emph{dead-end heuristic} \cite{bjornsson06} and \emph{Swamps-based
pathfinding} \cite{pochter10} are two closely-related pruning techniques
that identify areas in the search space not relevant for reaching the goal. 
This is a similar yet complementary goal
to RSR, which tries to reduce the search effort involved in exploring any given
area. Both methods decompose the map into a
series of obstacle-free areas. A preliminary
online search in the decomposed graph is then used to identify areas that can 
be ignored during a subsequent search in the original grid.
\par
The \emph{gateway heuristic} \cite{bjornsson06} and the \emph{portal heuristic}
\cite{goldenberg10} are two similar memory-based techniques which also attempt
to speed up optimal pathfinding on grid maps.  
Both decompose the map into a series of adjacent areas and both
pre-compute a database of exact distances between all pairs of nodes that transition 
from one area to another (called variously ``'portals'' or ``gates'').  
The main idea is to use this information to improve the accuracy of cost-to-go estimates
during search as a way of reducing the number states expanded by A*. 
Where the portal heuristic differs from other similar works
\cite{bjornsson06,sturtevant09} is in its use of the decomposed
graph to further prune the state space. A preliminary search 
identifies portals relevant to the problem instance at hand and, during 
a subsequent search in the original grid, 
any nodes from an area that contains no relevant portals are ignored.
\par
In the algorithm engineering community the problem of quickly computing optimal
shortest paths has received significant attention.  State of the art methods
such as Contraction Hierarchies \cite{geisberger08}
are based on a combination of Dijkstra's algorithm together with
memory-intensive abstractions. 
Such algorithms are very fast but they are also highly optimised for road
networks in which certain topological properties hold true: for example, the
existence of ``highway'' edges that appear on most shortest paths between
arbitrary pairs of nodes.  Though mostly orthogonal to RSR, there has been very
little work applying these ideas to searching on grid maps. One recent result
however \cite{sturtevant10} suggests they are not as effective when the
underlying graph contains a high degree of path symmetry.

%% file: empty_rooms.tex
\section{Rectangular Symmetry Reduction}
\label{sec:rsr}

We begin by making precise the notion of a symmetric relationship between paths
in a uniform cost graph:
\begin{definition}
\label{def:symmetry}
Two paths $\pi_{1}$ and $\pi_{2}$ are symmetric if they share the same start and
goal node and one can be derived from the other by interchanging the order of the
moves.
\end{definition}

When applying Definition \ref{def:symmetry} to an undirected uniform cost grid 
\footnote{We say uniform but infact straight moves cost 1 and diagonal
moves cost approx. $\sqrt{2}$.} we notice that each node expanded can
often be reached from one or more of its ancestors in the search tree by several 
symmetric paths of equal length.
If the nodes on these alternative paths have an $f$-value smaller than the
current node (even an equal $f$-value is often sufficient), A* will needlessly
expand them.  
We address this problem using the high-level strategy in Algorithm
\ref{alg:rsr}; which identifies and eliminates path symmetries from the grid.

\input alg_rsr

Our approach has similarities with 4ERR~\cite{harabor10}, a symmetry breaking algorithm 
limited to 4-connected grid maps.
The main differences are: (i) we generalise 4ERR to 8-connected grid maps 
(ii) we give a stronger offline pruning operator to eliminate more nodes from
the grid (iii) we give a new online pruning operator that reduces a node's branching
factor and further speeds up search.

The generalisation to uniform-cost 8-connected grids is more challenging than it might look
at a first glance.  On a 4-connected map no node requires more than one
macro-edge (to the closest node on the opposite side of the perimeter) to retain
optimality~\cite{harabor10}.  Thus, it is easy to maintain a low branching
factor.
As we show in the next section, many more macro-edges are
needed to preserve optimality on 8-connected maps. We will identify a set of
macro-edges that is necessary and sufficient to ensure that empty rectangles can
be crossed optimally.  Keeping the branching factor within reasonable limits is
a primary motivation for the enhancements reported in the next sections.

%% file: alg_rsr.tex
\begin{algorithm}
\caption{Graph reduction based on empty rectangles}
\label{alg:rsr}
\begin{algorithmic}
\REQUIRE A grid map 
\begin{enumerate}
\item{\label{alg:rsr:1} Decompose the grid map into a series of disjoint rectangles that are free of any obstacles.
As per \cite{harabor10}, the size and placement of the rectangles can vary across a
map, depending on the positions of obstacles.}
\item{\label{alg:rsr:2} Prune all tiles from the interior of each rectangle $R$
and possibly some from the perimeter (border).}
%, and any tiles from the
%perimeter which have no neighbours in any adjacent room.} 
\item{\label{alg:rsr:3} Add to each rectangle $R$ \emph{macro edges} between 
selected pairs of tiles from the perimeter. The cost of each edge is equal to
the Octile (or Manhattan, on 4-connected grids) distance between endpoints.}
\item{\label{alg:rsr:4} During search, temporarily re-insert tiles back into the map to handle cases where the
start or goal is a location which has been previously pruned.}
\end{enumerate}
\end{algorithmic}
\end{algorithm}

%% file: eight_connected.tex
\section{Optimal Room Traversal with Macro-Edges}

After interior nodes are eliminated, macro-edges between selected pairs of
perimeter nodes have to be added to ensure that rectangles can be traversed
optimally.  A straightforward approach would be adding a macro-edge between any
two nodes on the perimeter of a rectangle.  We will call the subgraph resulting
from such an operation a \emph{perimeter clique}.

Although the perimeter clique approach guarantees optimality, it has the
disadvantage of creating up a large branching factor and slowing down search
(the number of necessary macro edges is quadratic in the number of 
perimeter nodes).  We
introduce an alternative strategy, that creates much fewer macro-edges, by
defining a \emph{dominance} relation between macro-edges.

\begin{definition}
\label{def:dominance}
A macro-edge connecting two arbitrary nodes $t_1$ and $t_2$ in a perimeter
clique is non-dominated if all other paths between $t_1$ and $t_2$ in the
perimeter clique have a cost strictly larger than the macro-edge at hand.
\end{definition}

By starting with a perimeter clique and applying Definition \ref{def:dominance}
it is easy to see that the set of non-dominated macro-edges are precisely the
ones that we identify below.  There are three cases to discuss: connections
between nodes on the same rectangle side, connections between orthogonal
rectangle sides, and connections between opposite rectangle sides.
In the discussion that follows note that the length of each added macro-edge 
is equal to the heuristic distance between its two endpoints -- as measured
using Octile distance.

\begin{figure}[tb]
       \begin{center}
		   \includegraphics[width=0.97\columnwidth, trim = 10mm 10mm 10mm 0mm]
			{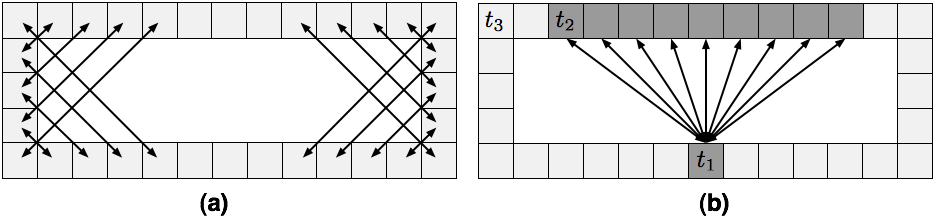}
       \end{center}
	\vspace{-3pt}
       \caption{(Left) Macro edges between nodes on orthogonal sides of an empty
       rectangle. (Right) Each node on the perimeter is connected to a set of 
		nodes on the opposite side.}
       \label{fig-macroedges}
\end{figure}

The first case is simple: adjacent nodes on the same perimeter side are connected
just as in the original grid. 
In the second case, two nodes on orthogonal sides of the perimeter of a rectangle
$R$ are connected \emph{iff} the shortest path between them is a diagonal
(45-degree) line; this is illustrated in Figure \ref{fig-macroedges} (left).
Notice that in both cases we introduce no more than two macro edges per node.
In the third case, we generate for each perimeter node a ``fan'' of neighbours
from the opposite side $R$.  Figure \ref{fig-macroedges} (right), illustrates
this idea.  Starting from a node such as $t_{1}$ we step to the closest
neighbour from the opposite side of $R$ and extend the fan by progressing away
from the middle in both directions adding each node we encounter.  The last node
on either side of the fan is placed diagonally, at 45 degrees, from $t_{1}$
(such as $t_{2}$) or located in the corner of the perimeter (whichever we
encounter first).  There is no need to add further nodes, such as $t_{3}$, as
these can be reached optimally from $t_1$ via the path $\lbrace t_1, t_2, \dots,
t_3\rbrace$.

In the rest of this paper, by macro-edge we will mean a non-dominated macro-edge.
We show next that the non-dominated macro-edges computed using our strategy are both
necessary and sufficient to ensure optimal traversal between any two perimeter
nodes.

\begin{proposition} All non-dominated macro-edges are necessary to ensure
optimal paths in a perimeter clique.  
\end{proposition} 
\begin{proof} By
definition, a non-dominated macro-edge $e$ is the only way to travel optimally
between its end nodes in a perimeter clique. Therefore, dropping $e$ would
result in losing path optimality in the perimeter clique.  
\end{proof}

\begin{lemma} \label{lemma-rooms} Let $R$ be an empty rectangle in
an 8-connected grid map. Let $m$ and $n$ be two perimeter locations.
Then, $m$ and $n$ can be connected optimally through a path that
contains only non-dominated macro-edges.
\end{lemma}

\begin{proof}Sketch:
We split the proof over the 3 cases discussed earlier: 1) {$m$ and $n$ are on
the same side of the perimeter;} 2) {\label{lemma-rooms-step2}$m$ and $n$ are on
orthogonal sides of the perimeter;} and 3) {\label{lemma-rooms-step3} $m$ and
$n$ are on opposite sides of the perimeter.}

In the first case we can simply walk along the perimeter from $m$ to $n$; the
optimality of this path is immediate. In the second and third case we argue as
follows:
the two nodes can be connected through an optimal path that has one diagonal macro-edge
(at one end of the path) and zero or more straight macro-edges.
See again the example of travelling from $t_1$ to $t_3$ in Figure
\ref{fig-macroedges} (Right).
\end{proof}

\noindent
\textbf{Node Insertion:}
Sometimes a node from the interior of an empty rectangle is required as a start
or goal location for an agent.  To handle such situations we give an online
procedure that temporarily re-inserts nodes back into the map for the duration
of a search.  It proceeds as follows: {If the start and goal are interior nodes
in the same room no insertion is necessary; an optimal path is trivially
available. } {On the other hand, if the start and goal are not in the same
rectangle, add four ``fans'' (collections) of macro edges.  Each fan connects
the start (goal) node to a set of nodes on one side of the rectangle's
perimeter.  Fans are built as shown earlier.}

Given the simple geometry of rectangles, it is possible to identify in constant
time the set of nodes which the start or goal must be connected to.  Further,
these neighbours could be generated on the fly.

\begin{lemma}
\label{lemma-insertion}
Let $R$ be an empty rectangle in an 8-connected grid map.  For any
nodes $m$, $n$, with $m$ a re-inserted interior node and $n$ a node on the
perimeter, it is always possible to find an optimal length path which mentions
no interior nodes except for $m$.
\end{lemma}
\begin{proof}
Our re-insertion procedure connects the start or goal to a set of nodes on each
side of $R$.  The procedure in each case is the same as the one given when
connecting two nodes on opposite sides of $R$.  To prove optimality we can
simply run the argument given for Step 3 of Lemma \ref{lemma-rooms} for each
node on the perimeter of $R$, in each case substituting $m$ for the newly
inserted node.
\end{proof}

We claim that eliminating symmetries as outlined earlier
preserves the completeness and the solution optimality:
\begin{theorem}
For every optimal path $\pi$ on an original grid, there exists an optimal path
$\pi'$ on the modified graph with the property that $\pi$ and $\pi'$ have the
same cost.
\end{theorem}
\begin{proof}
Consider an optimal path $\pi$ on the original map and a rectangle $R$ that is
crossed by $\pi$.  Let $m$ and $n$ be the two perimeter points along $\pi$.
According to Lemma~\ref{lemma-rooms}, there is a way to connect $m$ and $n$
optimally in the modified graph. Thus, we can replace the original segment $[m
\dots n]$ in $\pi$ with the cost-wise equivalent segment that corresponds to the
modified graph.  The case when $m$ (or $n$) is the start or goal node is
addressed similarly using Lemma~\ref{lemma-insertion}.  By performing such a
path segment replacement for all rectangles intersected by $\pi$, we obtain a
path $\pi'$ that satisfies the desired properties.
\end{proof}

%% file: branching_factor.tex
\section{Reducing The Branching Factor Further}
Consider an empty rectangle of width $w$ and height $h$ where $w > h$.  After
adding all non-dominated macro edges, each node from the perimeter will have
between $h$ to $2h-1$ neighbours from the opposite side of the rectangle, up to
2 neighbours from the same side of the rectangle and up to 5 other neighbours
from adjacent rectangles.  Such a high branching factor is undesirable as
individual node expansion operations take longer.

In this section we study two branching factor reduction methods.  The first is
an offline technique that prunes nodes from the perimeter of each rectangle.
The second is an online pruning strategy which we apply during individual node
expansion operations.  We discuss both methods in the context of 8-connected
grid maps however they are equally applicable to 4-connected maps.

\begin{figure}[t]
	\begin{center}
	\includegraphics[width=0.97\columnwidth, trim = 10mm 10mm 10mm 0mm]
	{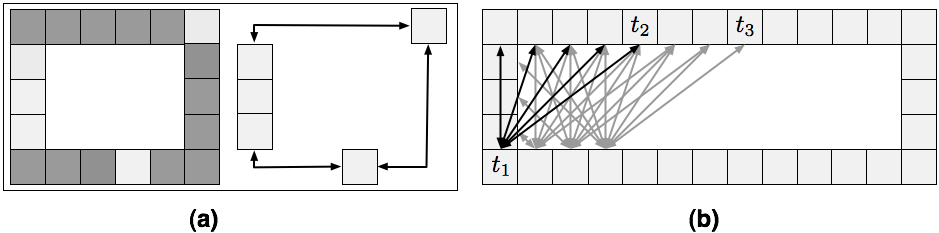}
	\end{center}
	\vspace{-3pt}
	\caption{(Left) From each empty rectangle we prune all (dark grey) nodes which
	have no neighbours in any adjacent rectangle.
	Remaining nodes are then connected directly.
	(Right) Assume that $t_{1}$ is the parent of $t_2$. When $t_2$
	is expanded, there is no need to generate its secondary neighbors.
	These can be reached directly from $t_1$ on a shorter or equal-length path.
}
\label{fig-branching}
\end{figure}

\noindent
\textbf{Perimeter Reduction:}
We observe that in many cases there are nodes on the perimeter of an empty
rectangle which have no neighbours from any adjacent rectangle.  These nodes
represent intermediate locations between entry and exit points that lead into
and out of each empty rectangle. To speed up search we propose pruning from the
perimeter of each rectangle all such nodes.  To preserve optimality, we will
connect the neighbours of each pruned node directly to each other.  The weight
of each new edge is set appropriately to the octile distance between the two
neighbours.  Figure \ref{fig-branching} (Left) shows an example.  As we will see
this optimisation can have a dramatic effect on the average performance of A* on
certain types of maps.

\begin{lemma}
Perimeter reduction preserves path optimality.
\end{lemma}
\begin{proof}
Sketch: Each time we prune a node from the perimeter we add a new edge between
all its neighbours with weight equal to the distance between each pair
of neighbours.  Thus, if a path exists between a pair of nodes before the
application of perimeter reduction it is guaranteed to exist afterward.
Further, the length of this path is unchanged.
\end{proof}

\noindent
\textbf{Online Node Pruning:}
Given a perimeter node $n$, let us partition its macro neighbors 
(connected to $n$ by macro-edges)
on the perimeter into 
\emph{primary neighbours} and \emph{secondary neighbours}.  Secondary neighbours are those
which are located on the opposite side of the perimeter to as compared to $n$
(excluding any corner nodes).  Primary neighbours are all the rest.

When expanding an arbitrary node from the perimeter of a rectangle we observe
that it is not necessary to consider any secondary neighbours if both the node
and its predecessor belong to the same rectangle. Figure \ref{fig-branching}
(Right) shows an example of such a situation; any path to a secondary neighbour
is strictly dominated by an alternative path through the predecessor. 
We apply this observation as follows: {During
node expansion, determine which rectangle the parent of the current node belongs
to.} {If the current node has no parent or the parent belongs to a different
rectangle, then process (i.e., generate) all primary and secondary neighbours.
Otherwise, process only primary neighbours.}

\begin{lemma}
Online node pruning preserves path optimality.
\end{lemma}
\begin{proof}
Sketch: Let $m$ be a node on the perimeter of a rectangle. Assume that its
parent $p$ belongs to the same rectangle.  Let $n$ be a secondary successor of
$m$.  Recall that $n$ and $m$ are on opposite sides of the rectangle.  We argue
below that passing through $m$ cannot possibly improve the best path between $p$
and $n$.  Therefore, there is no need to consider $(m,n)$ macro-edges when $m$
and $p$ belong to the same rectangle.

There are 4 cases when a node $m$ and its parent $p$ belong to the same
rectangle. In case 1, $p$ is a re-inserted node from the interior of the
rectangle.  Obviously, the path segment $p,m,n$ is suboptimal, as we zigzag from
$p$ to $m$ on one side of the rectangle and then to $n$ on the opposite side of
the rectangle.  In cases 2, 3, and 4, $p$ and $m$ are on opposite sides, on
orthogonal sides or on the same side of the rectangle. As in case 1, it is
possible to check in each case that taking a detour through $m$ does not improve
the shortest path from $p$ to $n$.
\end{proof}

%% file: memory.tex
\section{Memory Requirements and Dynamic Environments}
\label{sec:memory}
\textbf{Memory Requirements: }
In the most straightforward implementation, RSR requires storing the id of the
parent rectangle for each node in the original grid. 
This equates to an $O(|V|)$ memory overhead, where $V$ is the set of nodes
in the underlying graph.
Due to the simple geometric nature of empty rectangles, the set of macro
edges associated with each perimeter node can be computed on-the-fly in
constant time. 
\newline \noindent
\textbf{Dynamic Environments: }
In many application areas, most notably video games, the assumption of a static 
environment is sometimes unreasonable.
For example: obstacles may appear on the grid or existing obstacles may be
destroyed as the game progresses.
In such cases the underlying graph representing the world must be updated.
If a new obstacle appears, or an existing one is destroyed,
we can simply invalidate the affected rectangles and recompute new ones.
The repair operation must be very fast as most such applications run in real
time. As we will show, RSR appears particularly well suited for this task.

%% file: setup.tex
\section{Experimental Setup}
We evaluate the performance of RSR on three benchmarks taken from the freely
available pathfinding library Hierarchical Open Graph
(HOG)\footnote{\url{http://www.googlecode.com/p/hog2}}: {\textbf{Adaptive Depth}
is a set of 12 maps of size 100$\times$100 in which approximately $\frac{1}{3}$
of each map is divided into rectangular rooms of varying size and a large
open area interspersed with large randomly placed obstacles.} {\textbf{Baldur's
Gate} is a set of 120 maps taken from BioWare's popular roleplaying game
\emph{Baldur's Gate II: Shadows of Amn}.  Often appearing as a standard
benchmark in the literature \cite{bjornsson06,harabor10,pochter10} these maps
range in size from 50$\times$50 to 320$\times$320 and have a distinctive
45-degree orientation.} {\textbf{Rooms} is a set of 300 maps of size
256$\times$256 which are divided into symmetric rows of small rectangular areas
($7\times7$), connected by randomly placed entrances. This benchmark has
previously appeared in \cite{sturtevant09,pochter10,goldenberg10}.}
As discussed later, we also use a variant of each benchmark where every map is
scaled up by a factor of 3. In effect, our input data contains 864 maps in
total, with sizes up to $960\times960$.  
\par
Since our work is applicable to both 4
and 8 connected grid maps we used two copies each map: one in which diagonal
transitions are allowed and another in which they are not.  
For each map we generated 100 valid problem instances, checking that every
instance could be solved both with and without the use of diagonal transitions.
Our test machine had a 2.93GHz Intel Core 2 Duo processor, 4GB RAM and ran OSX
10.6.2.  Our implementation of A* is based on one provided in HOG, which we
adapted to facilitate our online node pruning enhancement.

%% file: results.tex
\section{Results}
\label{sec-results}
To evaluate RSR we use a generic implementation of A* and discuss performance 
in terms of search time speedup. That is, the relative improvement to the average 
time A* needs to solve an instance when running on a pruned  vs. unpruned grid.
For example, a speedup of 2.0 is twice as fast (higher is better).
Note that on approximately 2\% of all instances the start and goal are located
in the same rectangle and RSR computes the optimal solution without
search.  We exclude these instances from our results on the basis that they are 
outliers, even though RSR solves them in constant time.

\par 
\begin{figure*}[t]
       \begin{center}
                       \includegraphics[width=0.97\columnwidth, trim = 10mm 10mm 10mm 0mm]{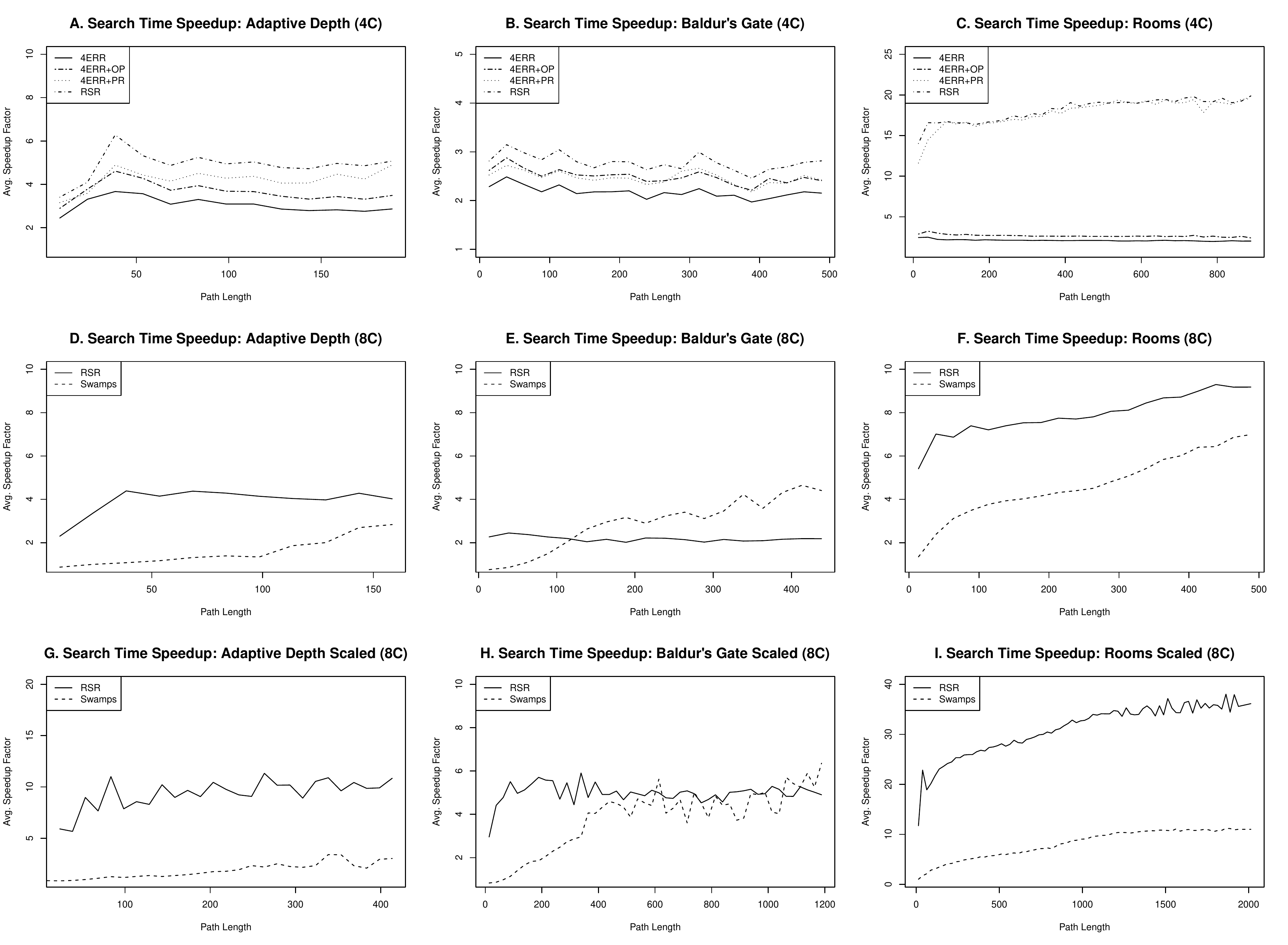}
       \end{center}
       \caption{Average A* speedup on each of our three benchmarks. 
		Results are given in terms of relative improvement to A* search time (i.e. speedup).}
\label{fig-speedup}
\end{figure*}

\textbf{Pre-processing Times: } 
Table 1 presents a summary of average pre-processing times for
each of our three (non-scaled) benchmarks. We also give the average number of
nodes and edges as an indication of map size.
We notice that RSR takes very little time to pre-process all input maps. 
We did not encounter any that took longer than a second, and most required 
significantly less than that. 
An interesting implication from this result is that RSR appears well suited to
pathfinding in dynamic environments as outlined in Section \ref{sec:memory}.

\input table_preproc

\textbf{Comparison to 4ERR: }
We now compare the performance RSR against the 4ERR~\cite{harabor10} graph
pruning algorithm.  As 4ERR works only on
4-connected grids, here we restrict our attention to this type of maps.  To
assess the individual impact of both perimeter reduction (PR) and online node
pruning (OP) we also develop and compare two variant algorithms: 4ERR+PR and
4ERR+OP. 
\par
Figure \ref{fig-speedup} (A to C) presents our main result.
Note that RSR shows a convincing 
speed improvement over 4ERR and all its variants across all input maps.
This allows us to conclude that that RSR is the better choice on 4-connected maps.
When analysing the impact of each enhancement, we note that 4ERR+PR yields the
biggest improvement on all three benchmarks, speeding up A* by up to 20 times.
4ERR+OP compares well with 4ERR+PR on both Adaptive Depth and
Baldur's Gate but is of little benefit on Rooms where perimeter pruning has
already reduced the branching factor.
\par
The large performance variation from one benchmark to another can be attributed
to how effectively we can decompose the map. A good decomposition forms large
rectangles with few perimeter nodes after pruning. This is the case for Rooms.
A poor decomposition builds small rectangles with many transitionary perimeter 
nodes that cannot be pruned. This is the case for Baldur's Gate. 
\par
\textbf{Comparison to Swamps:}
Next, we compare and contrast the performance of RSR with the Swamps
algorithm~\cite{pochter10}.  To evaluate Swamps we used the authors' source
code, including their own implementation of A*.  We then ran all experiments
using their recommended running parameters: a swamp seed radius of 6 and ``no
change limit'' of 2.
Figure \ref{fig-speedup} (D to F) gives search time speedup results for both RSR
and Swamps running on the 8-connected variants of our three benchmark problem
sets. 
On Adaptive Depth and Rooms, where the terrain can be naturally decomposed into
rectangles, RSR achieves higher speedups and is shown consistently better than Swamps. 
On Baldur's Gate, where this is not the case, Swamps-based pruning is more
effective. 
\par
Next, we scaled every map in each benchmark by a factor of 3 and randomly
generated a new set of 100 problem instances per map.  Scaling has the effect of
producing larger open areas and allows us to measure the impact of this variable
on search time speedup.  We present our findings in  Figure \ref{fig-speedup} (G
to I).  We observe that while the maximum speedup achieved by both algorithms
has increased, the gain for Swamps is very small while RSR shows dramatic
improvement.  Infact, if we limit our attention to problems of similar length to
those seen on the original maps we notice that the performance of Swamps
actually decreases.
\par
The observed performance characteristics are not unexpected: Swamps prune out
areas that can be avoided without introducing a detour while rectangle-based
symmetry reduction allows for a faster exploration of areas that need to be
searched.  Since it appears that the two algorithms are orthogonal, a natural
extension of this work would be to combine the two: first, apply 4(or 8)ERR+PR
(as appropriate) to a grid in order to eliminate as many interior nodes as
possible; then, apply a Swamps-based decomposition to the resultant graph.

\input table_portals

\textbf{Comparison to Portal Heuristic:}
We now compare RSR with PH-e -- the enhanced variant of the recent 
Portal Heuristic algorithm \cite{goldenberg10}.
Although we did not have access to a working implementation of this method we
will discuss its performance vs. RSR based on published results obtained by the
original authors. As in \cite{goldenberg10} we focus on the 4-connected variants 
of the Baldur's Gate and Rooms benchmarks.
Table 2 summarises the main result.
\par
PH-e performs well when it can decompose the map into areas of similar size with
few transitionary nodes connecting them.
RSR performs well when it can decompose the map into large rectangles with few
perimeter nodes.
On Rooms, both decomposition approaches are highly effective. 
On Baldur's Gate both are comparatively less effective.
Notice however that PH-e requires up to 7 times more memory than RSR to achieve
similar results.
As with Swamps, we believe PH-e is entirely orthogonal to RSR and the two can be 
easily combined. For example, PH-e could be used to more accurately guide search
on a map pruned by RSR. Alternatively, symmetry elimination could be used to 
speed up pathfinding between successive pairs of portals during PH-e's refinement phase.

%% file: table_preproc.tex
\begin{table}
\label{table:preproc}
\begin{center}
\begin{tabular}{|c|c|c|c|c|}
\hline
\textbf{Benchmark} & \textbf{Avg. Nodes} & \textbf{Avg. Edges} & \textbf{Preproc RSR}
& \textbf{Preproc Swamps}\\ \hline
Adaptive Depth & 8765 &  32773 & 0.10 & 5.06 \\ \hline
Baldur's Gate & 4507 & 16557 & 0.65 & 3.15 \\ \hline
Rooms & 51437 & 166417 & 0.39 & 16.9 \\ \hline
\end{tabular}
\end{center}
\caption{Input map size and average pre-processing times (seconds). }
\end{table}

%% file: table_portals.tex
\begin{table}[tb]
\begin{center}
\begin{tabular}{cccc}
\hline
\textbf{Algorithm} & \textbf{Extra Memory} & \textbf{Baldur's Gate} & \textbf{Rooms}  \\ \hline
PH-e & $2|V|$ & 3.16 &  11.9 \\ 
PH-e & $8|V|$ & 3.07 &  17.54 \\ 
RSR & $|V|$ & 2.8 & 18.2 \\ \hline
\end{tabular}
\end{center}
\caption{Avg. A* search time speedup: RSR vs PH-e. 
RSR figures are across all maps on each benchmark. 
PH-e figures are for a small subset selected by its authors (1 of 120 from
Baldur's Gate and 5 of 300 from Rooms). }
\label{table-phspeedup}
\end{table}

%% file: conclusion.tex
\section{Conclusion}
We introduce RSR, a new search space reduction algorithm applicable to
pathfinding on uniform cost grid maps. RSR is fast, memory efficient,
optimality preserving and can, in some cases, eliminate entirely the need
to search.  
When running on a grid pruned by RSR, A* is up to 38 times faster than
otherwise.
\par
We compare RSR with a range of search space reduction algorithms from the
literature. Compared to 4ERR~\cite{harabor10}, on which it is based, RSR is
shown significantly faster on the set of instances where both methods can be
applied (i.e. 4-connected maps).  Next, we compare RSR to Swamps-based
pruning~\cite{pochter10} and show that the two algorithms have complementary
strengths.  We find that Swamps are more useful on maps with small open areas
while RSR becomes more effective as larger open areas are available on a map. We
also identify a broad range of instances where RSR dominates convincingly and is
clearly the better choice.  Finally, we compare RSR to the enhanced Portal
Heuristic~\cite{goldenberg10}.  We show that our method exhibits similar or
improved performance but requires up to 7 times less memory.  As with Swamps, we
find that the two ideas are complementary and could be easily combined.

\par
Future work includes reducing the branching factor in RSR further through the 
development of better map decompositions and stronger online node pruning
strategies.
Another interesting topic is combining RSR with Swamps or the Portal Heuristic.

%% file: acknowledgements.tex
\section{Acknowledgements}
We would like to thank Alban Grastien and Patrik Haslum for providing feedback
on early drafts of this work. 
We also thank Ariel Felner and Meir Goldenberg for providing us with assistance
in comparing RSR with their enhanced Portal Heuristic algorithm.

NICTA is funded by the Australian Government as represented by the Department of 
Broadband, Communications and the Digital Economy and the Australian Research 
Council through the ICT Centre of Excellence program.